\let\mypdfximage\pdfximage
\def\pdfximage{\immediate\mypdfximage}
\documentclass[10pt,twocolumn,letterpaper]{article}

\usepackage{cvpr}
\usepackage{times}
\usepackage{epsfig}
\usepackage{graphicx}
\usepackage{amsmath}
\usepackage{amssymb}
\usepackage{booktabs,multirow,array}
\usepackage{xcolor}
\usepackage{epsfig,subfig}
\usepackage{graphicx}
\graphicspath{{examples/},{/imgs/}}
\usepackage{stfloats}

\usepackage{url}
\usepackage[pagebackref=true,breaklinks=true,letterpaper=true,colorlinks,bookmarks=false]{hyperref}

\cvprfinalcopy 
\ifcvprfinal\fi
\def\cvprPaperID{1503}

\def\imwh#1#2#3{\includegraphics[clip,width=#2\textwidth,height=#3\textheight]{#1.jpg}}
\def\bmat#1{\begin{bmatrix}#1\end{bmatrix}}

\def\fig#1{Fig.~\ref{fig:#1}}

\newcommand{\tb}[3]{\setlength{\tabcolsep}{#2mm}\begin{tabular}{#1}#3\end{tabular}}

\def\anp#1#2{{\it \textcolor{red}{#1} \textcolor{blue}{#2}}}

\def\anptable#1{
\begin{figure}[#1]
\centering

\setlength{\tabcolsep}{0.5mm}
\begin{tabular}{l|c|c|c|c}
\toprule[1pt]
\multirow{2}{*}{\bf\textcolor{red}{\tb{l}{0}{Adj-\\ective}}}&
\multicolumn{4}{c}{\bf\textcolor{blue}{Noun}}\\
\cline{2-5}
&

\it\textcolor{blue}{girls}& 
\it\textcolor{blue}{baby} & 
\it\textcolor{blue}{face} & 
\it\textcolor{blue}{eyes} \\
\midrule[1pt]
\rotatebox{70}{\it\textcolor{red}{adorable}} & 
\imwh{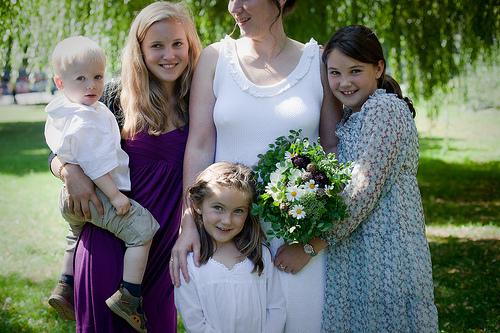}{0.1}{0.05}&
\imwh{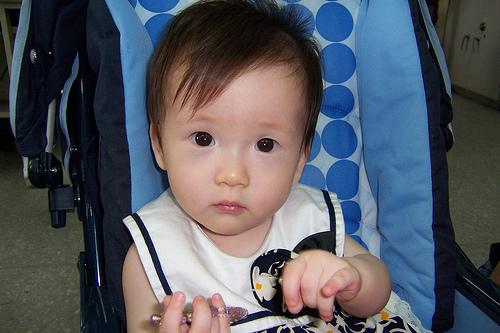}{0.1}{0.05}&
\imwh{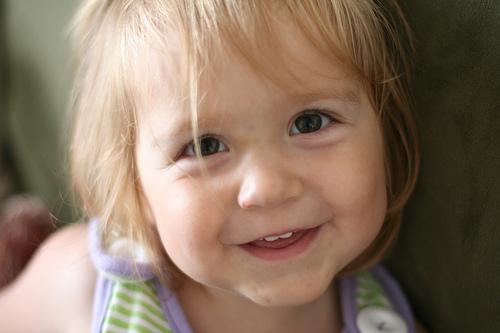}{0.1}{0.05}&
\\
&489&514&420&0\\
\midrule
\rotatebox{70}{\it\textcolor{red}{pretty}} & 
\imwh{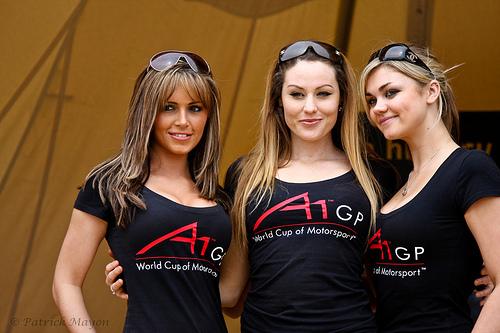}{0.1}{0.05}&
\imwh{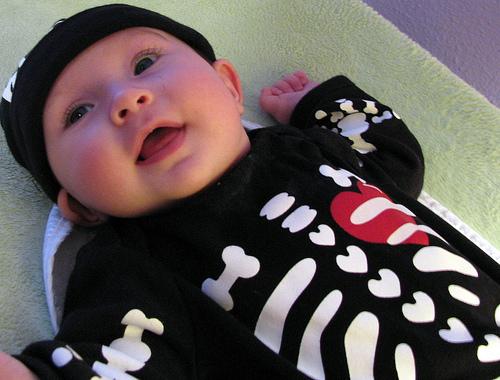}{0.1}{0.05}&
\imwh{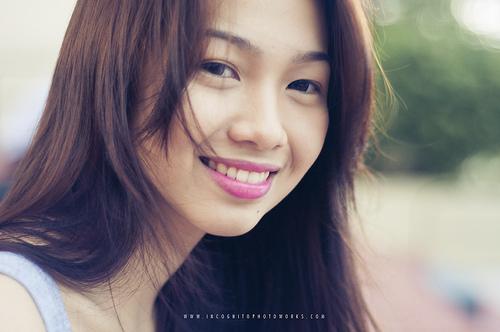}{0.1}{0.05}&
\imwh{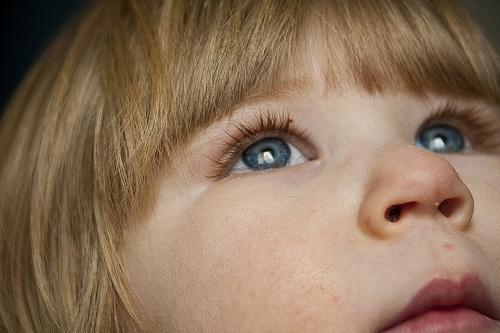}{0.1}{0.05}\\ 
&869 & 336 & 487& 703\\
\midrule
\rotatebox{70}{\it\textcolor{red}{attractive}} &
\imwh{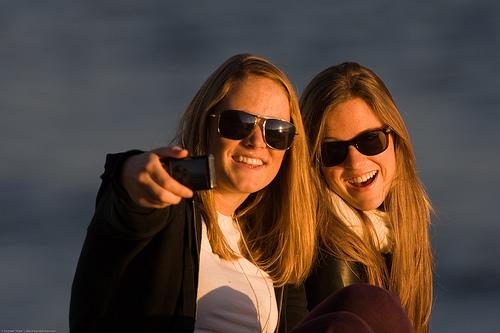}{0.1}{0.05}&
&
\imwh{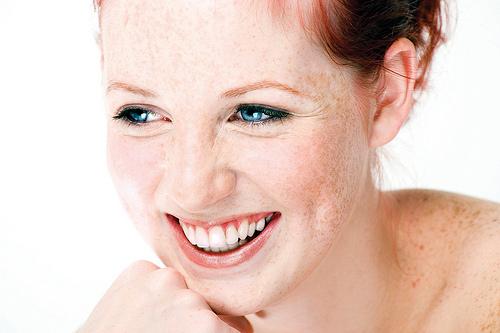}{0.1}{0.05}&
\imwh{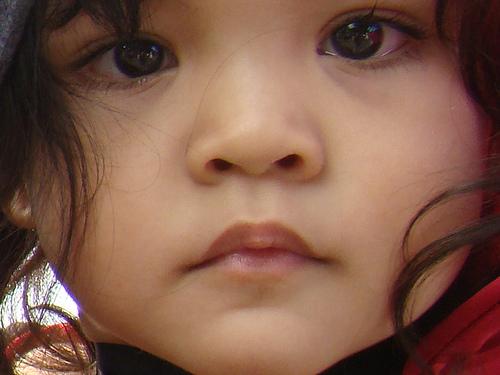}{0.1}{0.05}\\
&492& 0 & 430& 94\\
\bottomrule[1pt]

\end{tabular}

\caption{
Our task is to classify an image into Adjective Noun Pair (ANP) concepts. 
The numbers indicate the size of the ANP category in our dataset.
Our goal is to develop an ANP classifier out of extremely noisy training
data from the web that not only respects visual correlations along
adjectives (A) and nouns (N) semantics, but also fills in the semantic
blanks where there has been 0 training data.
}
\label{fig:anptable}
\end{figure}
}

\def\sampairLarge#1#2#3#4#5#6{
\tb{@{}cc@{}}{1}{
\tb{cccc}{0.1}{
\imwh{#101}{0.12}{0.05}&
\imwh{#102}{0.12}{0.05}&
\imwh{#103}{0.12}{0.05}&
\imwh{#104}{0.12}{0.05}\\[-2pt]
\imwh{#105}{0.12}{0.05}&
\imwh{#106}{0.12}{0.05}&
\imwh{#107}{0.12}{0.05}&
\imwh{#108}{0.12}{0.05}\\[-2pt]
\imwh{#109}{0.12}{0.05}&
\imwh{#110}{0.12}{0.05}&
\imwh{#111}{0.12}{0.05}&
\imwh{#112}{0.12}{0.05}\\
}
&
\tb{cccc}{0.2}{
\imwh{#201}{0.12}{0.05}&
\imwh{#202}{0.12}{0.05}&
\imwh{#203}{0.12}{0.05}&
\imwh{#204}{0.12}{0.05}\\[-2pt]
\imwh{#205}{0.12}{0.05}&
\imwh{#206}{0.12}{0.05}&
\imwh{#207}{0.12}{0.05}&
\imwh{#208}{0.12}{0.05}\\[-2pt]
\imwh{#209}{0.12}{0.05}&
\imwh{#210}{0.12}{0.05}&
\imwh{#211}{0.12}{0.05}&
\imwh{#212}{0.12}{0.05}\\
}\\
\it \textcolor{red}{#3} \textcolor{blue}{#4}&
\it\textcolor{red}{#5} \textcolor{blue}{#6}\\
}
}

\def\labelnoise#1{
\begin{figure*}[#1]
\centering
\sampairLarge{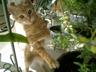}{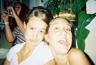}{pretty}{baby}{attractive}{face}
\caption{ User generated ANP tags of images are inherently noisy: The
  same noun ({\it \textcolor{blue}{baby}}) could mean different
  entities, and a positive adjective ({\it
    \textcolor{red}{attractive}}) could modify the pairing noun with a
  negative sentiment when used sarcastically.  }
\label{fig:labelnoise}
\end{figure*}
}

\def\overview#1{
\begin{figure*}[#1]
\centering\vspace{-5mm}
\includegraphics[clip,width=0.8\textwidth]{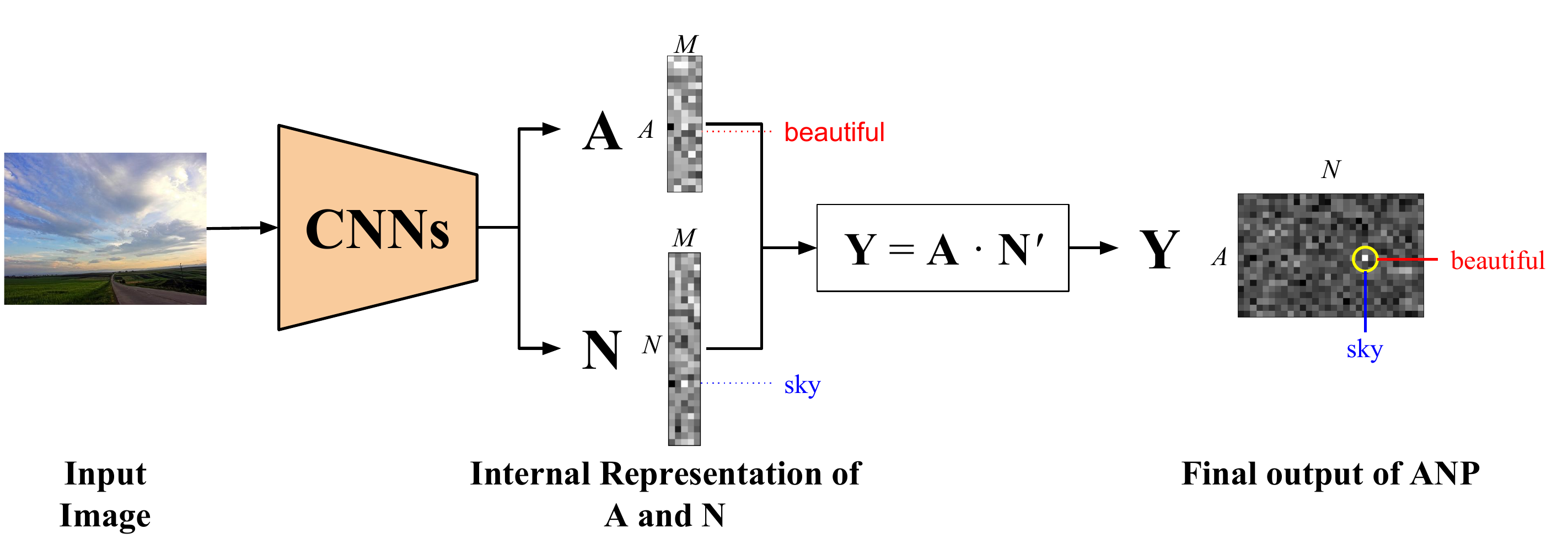}
\caption{
Overview of our factorized CNN model for ANP classification.
\label{fig:overview}}
\end{figure*}
}

\def\models#1{
\begin{figure*}[#1]
\centering
\tb{@{\hspace{-10mm}}c@{\hspace{-18mm}}c@{\hspace{-8mm}}c@{\hspace{-4mm}}cc}{1}{
\tb{c}{0}{\includegraphics[clip,width=0.28\textwidth]{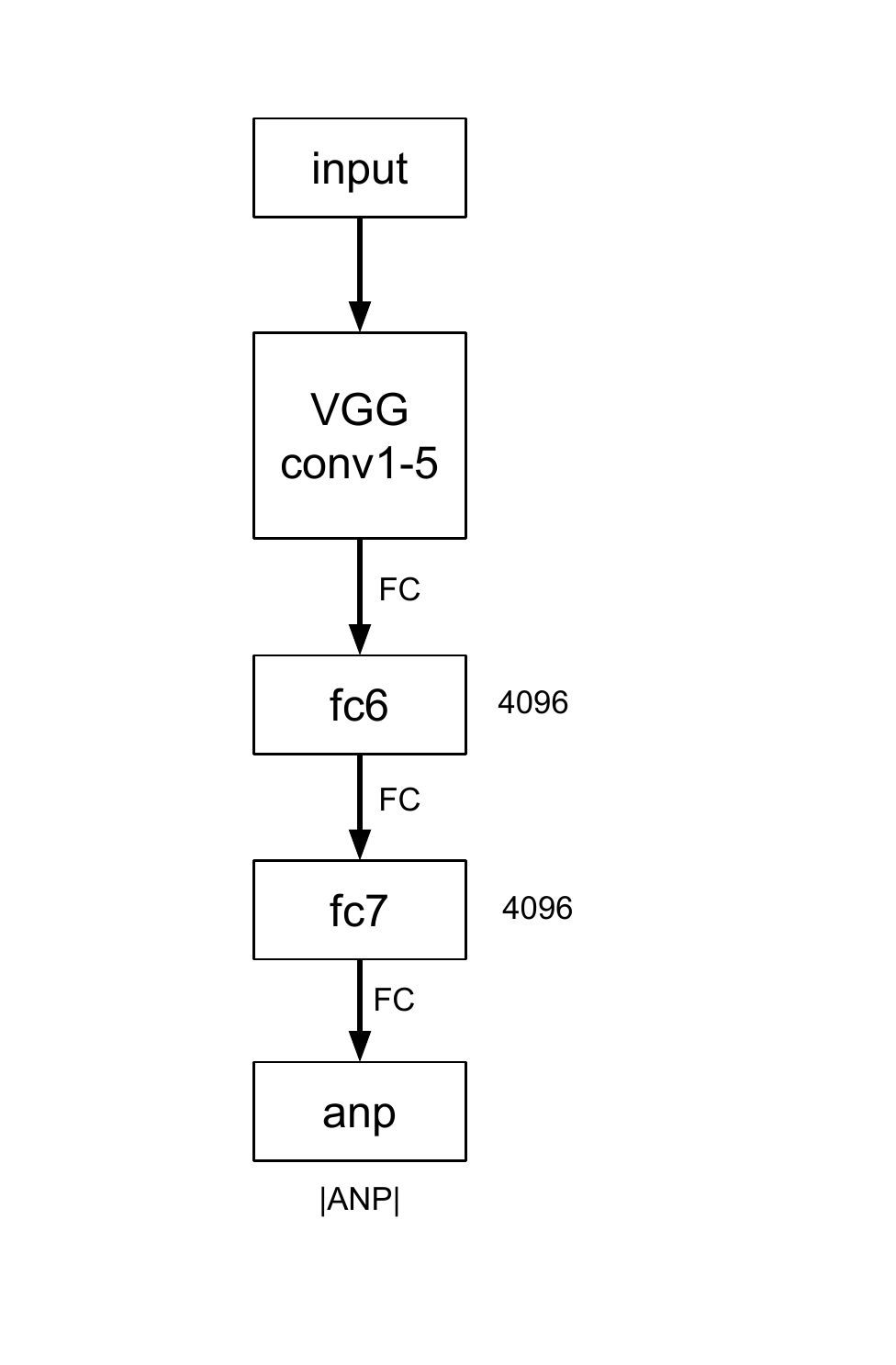}}&
\tb{c}{0}{\includegraphics[clip,width=0.28\textwidth]{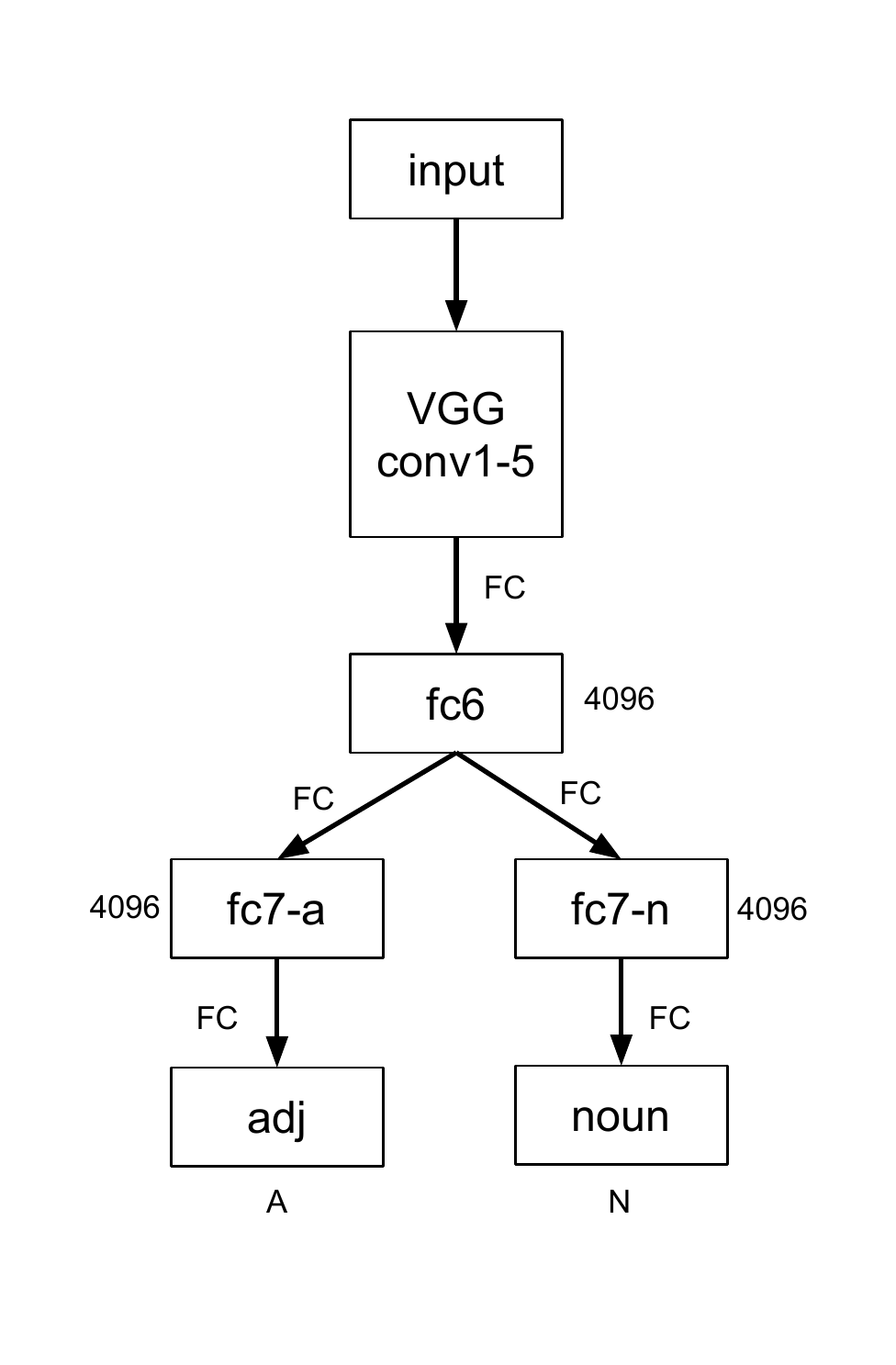}}&
\tb{c}{0}{\includegraphics[clip,width=0.28\textwidth]{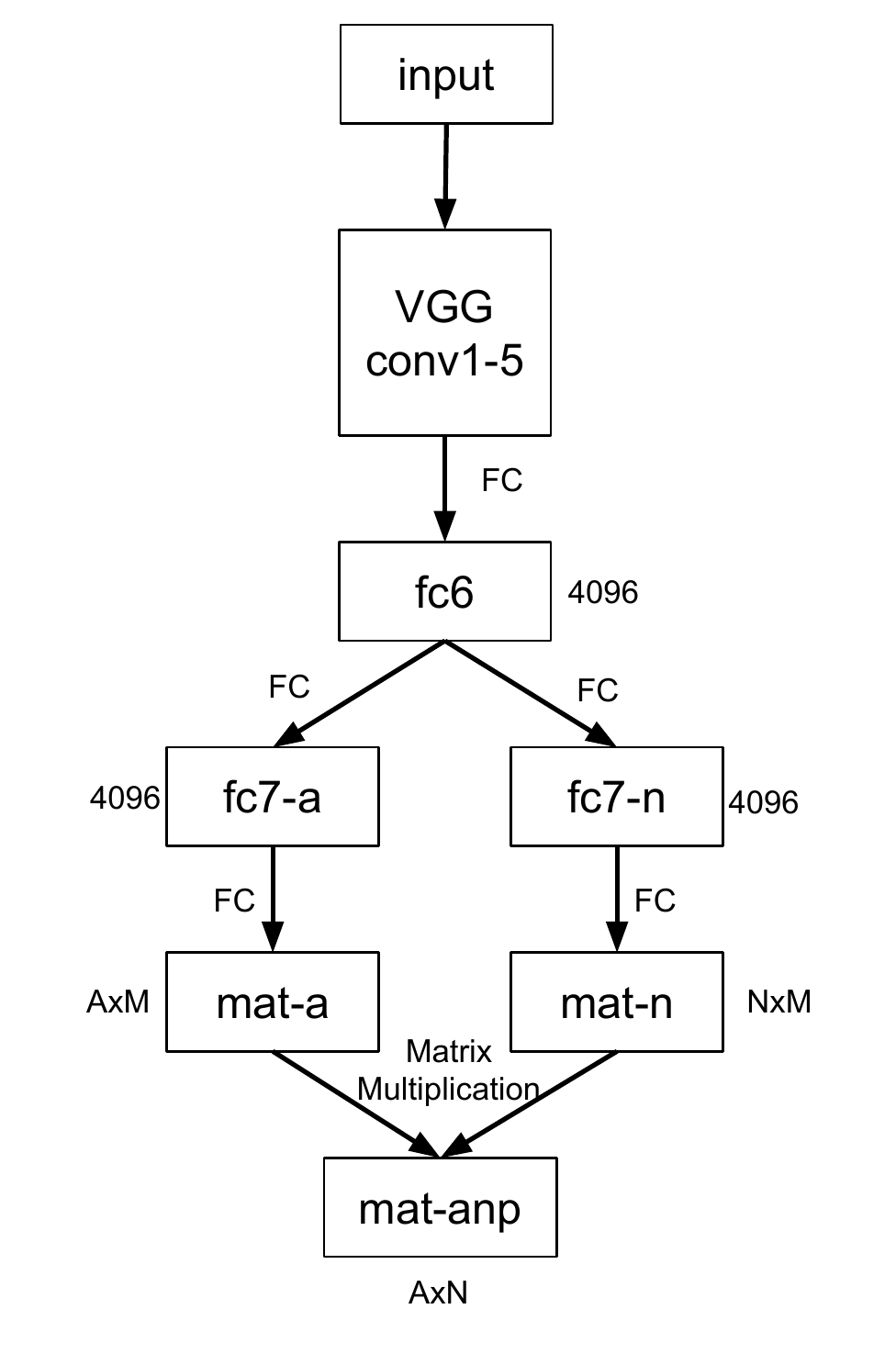}}&
\tb{c}{0}{\includegraphics[clip,width=0.2\textwidth]{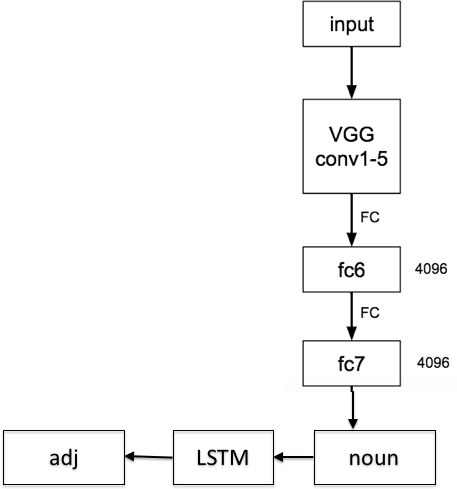}}&
\tb{c}{0}{\includegraphics[clip,width=0.2\textwidth]{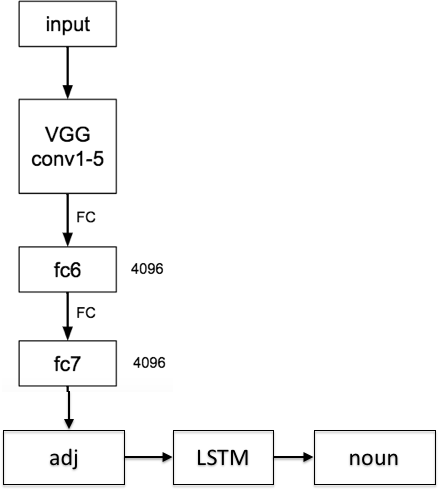}}\\
\bf a) ANP-Net&
\bf b) Fork-Net&
\bf c) Fact-Net&
\bf d) N-LSTM-A&
\bf e) A-LSTM-N\\
}
\caption{
Five deep convolutional neural network architectures used in our experiments. 
\label{fig:models}}
\end{figure*}
}

\def\vsotable#1{
\begin{table}[#1]
\caption{Visual Sentiment Ontology ANP statistics \cite{borth2013vso_sentibank}.\label{tab:retrieval_overview}
}
\scriptsize
\setlength{\tabcolsep}{1mm}
\begin{tabular}[b]{@{}lr@{}}

\setlength{\tabcolsep}{1mm}
\begin{tabular}[b]{@{}lrr@{}}
\toprule
& \textbf{Flickr} & \textbf{YouTube}\\ \midrule
emotions & 24 & 24 \\
images/videos & 150,034 & 166,342 \\
tags & 3,138,795 & 3,079,526 \\
distinct tags & 17,298 & 38,935 \\
tags per image & 20.92 & 18.51 \\
tag re-usage & 181.45 & 79.09 \\
distinct top 100 tags & \textbf{1,146} & \textbf{1,047} \\
\bottomrule
\end{tabular}
&

\setlength{\tabcolsep}{0.5mm}
\begin{tabular}[b]{@{}lr}
\toprule
\multicolumn{2}{r}{\textbf{VSO Statistics}}\\ \midrule
ANP candidates & {320,000} \\
ANP with images & 47,000 \\
ANPs in VSO & \textbf{3000} \\
\multicolumn{2}{l}{%
\begin{tabular}[b]{lr}
top ANP & 
\it \textcolor{red}{happy} \textcolor{blue}{birthday}\\
top positive A & \textcolor{red}{\it beautiful, amazing}\\%, cute}\\
top negative A & \textcolor{red}{\it sad, angry, dark} \\
top N & \textcolor{blue}{\it face, eyes, sky} \\
\end{tabular}
}\\
\bottomrule
\end{tabular}\\

\end{tabular}
\end{table}
}

\def\topkaccu#1{
\begin{table}[#1]   
\centering
\caption{
Top-$k$ accuracies (\%) over all ANPs.
}\label{tab:topk}

\begin{tabular}{@{}c@{}}

\setlength{\tabcolsep}{3mm}
\begin{tabular}{l|rrr}
\toprule
\bf a) Seen ANP & top-1 & top-5 & top-10 \\ \midrule
DeepSentiBank~\cite{chen2014deepsentibank} & 9.77 & 22.11 & 29.59 \\
ANP-Net & 13.56 & 30.61 & 40.16 \\
Fork-Net & 10.82 & 25.76 & 34.53 \\
Fact-Net $M=2$ & 12.51 & 28.95 & 38.32 \\
Fact-Net $M=5$ & 11.68 & 27.80 &  37.28 \\
Fact-Net $M=10$ & 13.22 & 30.36 & 39.92 \\
N-LSTM-A & 3.362&-&- \\
A-LSTM-N & 3.321&-&-  \\
chance & 0.066 &0.33& 0.66\\
\bottomrule
\end{tabular}
\\
\\
\setlength{\tabcolsep}{3mm}
\begin{tabular}{l|rrr}
\toprule
\bf b) Unseen ANP & top-1 & top-5 & top-10 \\ \midrule
DeepSentiBank~\cite{chen2014deepsentibank} & - & - & - \\
ANP-Net & - & - & -\\
Fork-Net & 0.27 & 3.35 & 7.09 \\%0.24 & 2.92 & 6.31 \\
Fact-Net $M=2$ & \bf 1.54 & \bf 7.17 & \bf 11.85 \\ %1.36 & \bf 6.38 & \bf 10.86  \\
Fact-Net $M=5$ & 0.77 & 4.52 & 8.77\\%0.74 & 4.11 & 8.09 \\
Fact-Net $M=10$ & 0.36 & 3.35 & 7.22\\%0.32 & 2.99 & 6.55 \\
N-LSTM-A & 0.012&-&- \\
A-LSTM-N & 0.009&-&-  \\
\bottomrule
\end{tabular}

\end{tabular}
\end{table}
}

\def\sampair#1#2#3#4#5#6{
\tb{@{}cc@{}}{0.2}{
\tb{cccc}{0.1}{
\imwh{#101}{0.06}{0.03}&
\imwh{#102}{0.06}{0.03}&
\imwh{#103}{0.06}{0.03}&
\imwh{#104}{0.06}{0.03}\\[-2pt]
\imwh{#105}{0.06}{0.03}&
\imwh{#106}{0.06}{0.03}&
\imwh{#107}{0.06}{0.03}&
\imwh{#108}{0.06}{0.03}\\[-2pt]
\imwh{#109}{0.06}{0.03}&
\imwh{#110}{0.06}{0.03}&
\imwh{#111}{0.06}{0.03}&
\imwh{#112}{0.06}{0.03}\\[-2pt]
}
&
\tb{cccc}{0.2}{
\imwh{#201}{0.06}{0.03}&
\imwh{#202}{0.06}{0.03}&
\imwh{#203}{0.06}{0.03}&
\imwh{#204}{0.06}{0.03}\\[-2pt]
\imwh{#205}{0.06}{0.03}&
\imwh{#206}{0.06}{0.03}&
\imwh{#207}{0.06}{0.03}&
\imwh{#208}{0.06}{0.03}\\[-2pt]
\imwh{#209}{0.06}{0.03}&
\imwh{#210}{0.06}{0.03}&
\imwh{#211}{0.06}{0.03}&
\imwh{#212}{0.06}{0.03}\\[-2pt]
}\\
\it \textcolor{red}{#3} \textcolor{blue}{#4}&
\it\textcolor{red}{#5} \textcolor{blue}{#6}\\[2pt]
}
}

\def\accugap#1{
\begin{figure*}[#1]   
\centering
\scriptsize
\tb{cc}{1}{
\bf 
\tb{r}{0}{
a) Seen ANPs sorted by top-10 accuracy gap between Fact-Net and ANP-Net
}
&
\bf\tb{r}{0}{
b) Unseen ANPs sorted by top-10 accuracy gap between Fact-Net and Fork-Net
}
\\

\setlength{\tabcolsep}{0.5mm}
\begin{tabular}{@{\hspace{-4mm}}c|rrrrrccr@{}}
\toprule
\bf seen ANP &
\bf \tb{r}{0}{top-10\\Fact-Net} &
\bf \tb{r}{0}{top-10\\ANP-Net} &
\bf \tb{r}{0}{test\\size} &
\bf \tb{r}{0}{\#A train\\ images} &
\bf \tb{r}{0}{\#N train\\ images} &
\bf \tb{r}{0}{\#A train\\ ANPs} &
\bf \tb{r}{0}{\#N train\\ ANPs}&
\bf \tb{r}{0}{accu.\\gap}\\ 
\midrule
\anp{cute}{bird} & 0.600 & 0.378 & 45 & 5394 & 4214 & 16 & 12 & 0.222\\
\anp{sexy}{legs} &  0.647 & 0.482 & 85 & 4330 & 3422 & 13 & 10 & 0.165\\
\anp{curious}{dog} & 0.510 & 0.346 & 104 & 3285 & 18289 & 8 & 43 & 0.163\\
\anp{scary}{eyes} & 0.308 & 0.154 & 52 & 4970 & 10138 & 14 & 24 & 0.154\\
\anp{quiet}{lake} & 0.632 & 0.484 & 95 & 4352 & 6394 & 12 & 14 & 0.147\\
\anp{tasty}{food} & 0.455 & 0.309 & 110 & 1872 & 10372 & 6 & 24 & 0.145\\
\anp{serene}{lake} & 0.436 & 0.299 & 117 & 1430 & 6394 & 4 & 14 & 0.137\\
\anp{traditional}{food} & 0.662 & 0.527 & 148 & 6976 & 10372 & 14 & 24 & 0.135\\
\anp{grumpy}{baby} & 0.433 & 0.300 & 90 & 1389 & 9557 & 4 & 26 & 0.133\\
\anp{calm}{lake} & 0.708 & 0.575 & 106 & 4121 & 6394 & 12 & 14 & 0.132\\
\midrule
\anp{crazy}{girls} & 0.271 & 0.426 & 129 & 7412 & 11967 & 16 & 30 & -0.155\\
\anp{powerful}{animal} & 0.259 & 0.414 & 58 & 1253 & 1862 & 4 & 7 & -0.155\\
\anp{strange}{house} & 0.2 & 0.364 & 110 & 5062 & 9343 & 16 & 20 & -0.164\\
\anp{dry}{eye} & 0.138 & 0.303 & 109 & 7053 & 1474 & 19 & 4 & -0.165\\
\anp{little}{beauty} & 0.038 & 0.212 & 104 & 11241 & 5106 & 19 & 16 & -0.173\\
\anp{tough}{face} & 0.014 & 0.188 & 69 & 1240 & 18432 & 5 & 48 & -0.174\\
\anp{shy}{dog} & 0.278 & 0.454 & 97 & 2583 & 18289 & 7 & 43 & -0.175\\
\anp{super}{kids} & 0.242 & 0.424 & 99 & 3570 & 4761 & 11 & 13 & -0.182\\
\anp{dry}{river} & 0.384 & 0.592 & 125 & 7053 & 5488 & 19 & 14 & -0.208\\
\anp{dirty}{house} & 0.143 & 0.403 & 77 & 7174 & 9343 & 16 & 20 & -0.260\\
\bottomrule
\end{tabular}
&

\setlength{\tabcolsep}{0.5mm}
\begin{tabular}{@{\hspace{-4mm}}c|rrrrrccr@{}}
\toprule
\bf unseen ANP &
\bf \tb{r}{0}{top-10\\Fact-Net} &
\bf \tb{r}{0}{top-10\\Fork-Net} &
\bf \tb{r}{0}{test\\size} &
\bf \tb{r}{0}{\#A train\\ images} &
\bf \tb{r}{0}{\#N train\\ images} &
\bf \tb{r}{0}{\#A train\\ ANPs} &
\bf \tb{r}{0}{\#N train\\ ANPs}&
\bf \tb{r}{0}{accu.\\gap}\\ 
\midrule
\anp{cute}{cake} & 0.503 & 0.101 & 179 & 5394 & 2950 & 16 & 8 & 0.402\\
\anp{stormy}{field} & 0.472 & 0.111 & 144 & 1777 & 891 & 5 & 3 & 0.361\\
\anp{classic}{rose} & 0.569 & 0.215 & 181 & 2349 & 2411 & 7 & 6 & 0.354\\
\anp{smiling}{kids} & 0.377 & 0.068 & 146 & 2264 & 4761 & 7 & 13 & 0.308\\
\anp{magnificent}{butterfly} & 0.475 & 0.169 & 118 & 3577 & 866 & 12 & 3 & 0.305\\
\anp{shiny}{city} & 0.362 & 0.058 & 138 & 3633 & 8790 & 10 & 23 & 0.304\\
\anp{falling}{rain} & 0.342 & 0.043 & 161 & 1568 & 1677 & 4 & 4 & 0.298\\
\anp{cute}{toy} & 0.443 & 0.15 & 140 & 5394 & 733 & 16 & 2 & 0.293\\
\anp{nasty}{bathroom} & 0.353 & 0.065 & 139 & 821 & 749 & 3 & 2 & 0.288\\
\anp{rainy}{bridge} & 0.443 & 0.184 & 158 & 3604 & 2542 & 8 & 5 & 0.259\\
\midrule
\anp{safe}{car} & 0.007 & 0.082 & 147 & 747 & 6573 & 2 & 17 & -0.075\\
\anp{warm}{christmas} & 0.009 & 0.088 & 114 & 5663 & 1257 & 16 & 3 & -0.079\\
\anp{icy}{forest} & 0.149 & 0.255 & 141 & 3382 & 3989 & 8 & 11 & -0.106\\
\anp{successful}{student} & 0 & 0.129 & 140 & 797 & 1656 & 3 & 6 & -0.129\\
\anp{powerful}{river} & 0 & 0.149 & 134 & 1253 & 5488 & 4 & 14 & -0.149\\
\anp{sexy}{smile} & 0.034 & 0.203 & 177 & 4330 & 5338 & 13 & 16 & -0.169\\
\anp{derelict}{asylum} & 0.009 & 0.217 & 115 & 2807 & 961 & 7 & 2 & -0.209\\
\anp{salty}{waves} & 0.02 & 0.238 & 101 & 1024 & 2522 & 3 & 7 & -0.218\\
\anp{powerful}{ocean} & 0 & 0.221 & 122 & 1253 & 1372 & 4 & 4 & -0.221\\
\anp{derelict}{window} & 0 & 0.624 & 109 & 2807 & 2091 & 7 & 4 & -0.624\\
\bottomrule
\end{tabular}
\\
\\
\bf c) top \& bottom 3 seen ANP sample images & 
\bf d) top \& bottom 3 unseen ANP sample images\\ 

\tb{c}{0}{
\sampair{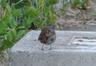}{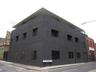}{cute}{bird}{dirty}{house}\\
\sampair{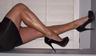}{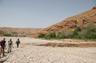}{sexy}{legs}{dry}{river}\\
\sampair{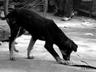}{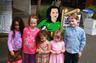}{curious}{dog}{super}{kids}\\
}
&
\tb{c}{0}{
\sampair{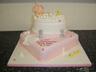}{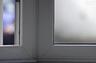}{cute}{cake}{derelict}{window}\\
\sampair{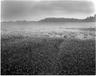}{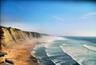}{stormy}{field}{powerful}{ocean}\\
\sampair{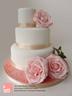}{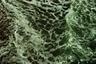}{classic}{rose}{salty}{waves}\\
}
\\
}
\caption{
Most and least accurate results by Fact-Net, 
compared to ANP-Net on seen ANPs and to Fork-Net on unseen ANPs.
\label{fig:accugap}
}
\end{figure*}
}

\def\sampset#1#2#3#4{
\tb{cccc}{0.2}{
\imwh{#101}{0.12}{0.05}&
\imwh{#102}{0.12}{0.05}&
\imwh{#103}{0.12}{0.05}&
\imwh{#104}{0.12}{0.05}\\[-2pt]
#2\\
\imwh{#105}{0.12}{0.05}&
\imwh{#106}{0.12}{0.05}&
\imwh{#107}{0.12}{0.05}&
\imwh{#108}{0.12}{0.05}\\[-2pt]
#3\\
\imwh{#109}{0.12}{0.05}&
\imwh{#110}{0.12}{0.05}&
\imwh{#111}{0.12}{0.05}&
\imwh{#112}{0.12}{0.05}\\[-2pt]
#4\\
}}

\def\retrievesource#1{
\begin{figure*}[#1]
\centering
\scriptsize
\tb{@{}c|c@{}}{0.5}{
\toprule
\bf images retrieved by Fact-Net &
\bf images retrieved by Fork-Net \\
\midrule
\multicolumn{2}{c}{\bf 1) seen ANP query: \anp{beautiful}{sky}} 
\\
\sampset{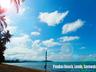}
{\anp{awesome}{trip}&\anp{beautiful}{sky}&\anp{bright}{sky}&\anp{clear}{lake}}
{\anp{incredible}{beauty}&\anp{beautiful}{sunset}&\anp{heavy}{clouds}&\anp{amazing}{sky}}
{\anp{harsh}{sea}&\anp{cloudy}{sunrise}&\anp{incredible}{sunset}&\anp{clear}{night}}
&
\sampset{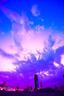}
{\anp{gentle}{flowers}&\anp{sunny}{sky}&\anp{beautiful}{clouds}&\anp{magnificent}{sky}}
{\anp{magical}{sunset}&\anp{clear}{night}&\anp{colorful}{sunset}&\anp{magical}{sunset}}
{\anp{colorful}{sky}&\anp{beautiful}{sky}&\anp{serene}{winter}&\anp{smooth}{clouds}}
\\ 
\midrule

\multicolumn{2}{c}{\bf 2) unseen ANP query: \anp{beautiful}{men}}
\\
\sampset{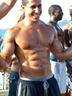}
{\anp{hot}{body}&\anp{hot}{model}&\anp{sexy}{fashion}&\anp{hot}{girls}}
{\anp{traditional}{dress}&\anp{handsome}{guy}&\anp{strong}{hair}&\anp{sexy}{body}}
{\anp{sexy}{body}&\anp{cold}{beer}&\anp{stupid}{graffiti}&\anp{smiling}{guy}}
&
\sampset{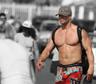}
{\anp{tough}{guy}&\anp{talented}{student}&\anp{relaxing}{bath}&\anp{sexy}{dance}}
{\anp{heavy}{book}&\anp{heavy}{weight}&\anp{hot}{butt}&\anp{incredible}{adventure}}
{\anp{strong}{men}&\anp{sexy}{fashion}&\anp{fragile}{body}&\anp{handsome}{face}}
\\ 
\midrule

\multicolumn{2}{c}{\bf 3) unseen ANP query: \anp{ugly}{baby}}
\\
\sampset{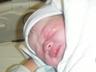}
{\anp{precious}{gift}&\anp{chubby}{face}&\anp{chubby}{baby}&\anp{chubby}{baby}}
{\anp{fresh}{baby}&\anp{warm}{hat}&\anp{laughing}{baby}&\anp{crying}{baby}}
{\anp{chubby}{baby}&\anp{funny}{face}&\anp{laughing}{baby}&\anp{fresh}{baby}}
&
\sampset{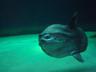}
{\anp{ugly}{fish}&\anp{favorite}{animal}&\anp{dead}{pig}&\anp{dangerous}{spider}}
{\anp{favorite}{animal}&\anp{ugly}{guy}&\anp{clean}{baby}&\anp{favorite}{animal}}
{\anp{graceful}{animals}&\anp{chubby}{face}&\anp{cute}{animals}&\anp{creepy}{doll}}
\\ 
\midrule

\multicolumn{2}{c}{\bf 4) unseen ANP query: \anp{ugly}{sky}}
\\
\sampset{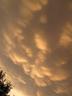}
{\anp{weird}{clouds}&\anp{weird}{clouds}&\anp{magical}{moon}&\anp{beautiful}{sky}}
{\anp{strange}{clouds}&\anp{amazing}{clouds}&\anp{fluffy}{bed}&\anp{amazing}{clouds}}
{\anp{weird}{clouds}&\anp{pleasant}{surprise}&\anp{beautiful}{sky}&\anp{weird}{clouds}}
&
\sampset{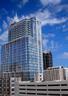}
{\anp{beautiful}{sky}&\anp{clear}{sky}&\anp{lovely}{city}&\anp{ugly}{building}}
{\anp{nice}{building}&\tiny\anp{magnificent}{architecture}&\anp{shiny}{city}&\tiny\anp{elegant}{architecture}}
{\anp{shiny}{gold}&\tiny\anp{amazing}{architecture}&\anp{nice}{building}&\anp{lovely}{evening}}
\\
\bottomrule
}

\caption{Images with their PGT tags for 1 seen and 3 unseen ANPs retrieved by Fact-Net and Fork-Net.
\label{fig:retrieve}
}
\end{figure*}
}

\title{
Mapping Images to Sentiment Adjective Noun Pairs with Factorized Neural Nets
}

\author{
Takuya Narihira\\
Sony / ICSI\\
\texttt{takuya.narihira@jp.sony.com} \\
\and
Damian Borth \\
DFKI / ICSI\\
\texttt{damian.borth@dfki.de} \\
\and
Stella X. Yu \\
UC Berkeley / ICSI \\
\texttt{stellayu@berkeley.edu} \\
\and
Karl Ni \\
In-Q-Tel \\
\texttt{kni@iqt.org} \\
\and
Trevor Darrell \\
UC Berkeley / ICSI\\
\texttt{trevor@berkeley.edu} \\
}

\begin{document}

\maketitle

\begin{abstract} 
  We consider the visual sentiment task of mapping an image to an adjective noun pair (ANP) such as "cute baby". To capture the two-factor structure of our ANP semantics as well as to overcome annotation noise and ambiguity, we propose a novel factorized CNN model which learns separate representations for adjectives and nouns but optimizes the classification performance over their product.  Our experiments on the publicly available SentiBank dataset show that our model significantly outperforms not only independent ANP classifiers on unseen ANPs and on retrieving images of novel ANPs, but also image captioning models 
  which capture word semantics from co-occurrence of natural text; the latter turn out to be surprisingly poor at capturing the sentiment evoked by pure visual experience.  That is, our factorized ANP CNN not only trains better from noisy labels, generalizes better to new images, but can also expands the ANP vocabulary on its own.
\end{abstract}

\section{Introduction}

\anptable{tp}

Automatic assessment of sentiment from visual content has gained considerable attention~\cite{borth2013vso_sentibank, chen2014deepsentibank, chen2014objectsentibank, xu2014visual, you2015robust}. One key element towards achieving this is the use of Adjective Noun Pair (ANP) concepts as a mid-level representation of visual content. 
We consider the task of labeling user-generated images by ANPs that visually convey a plausible sentiment, e.g. {\it adorable girls} in \fig{anptable}.  
This task can be more subjective and holistic, e.g. {\it beautiful
  landscape}~\cite{al2015makes}, as compared to object detection~\cite{li2010objectbank, russakovsky2014imagenet}, scene categorization~\cite{gong2014multi}, or pure visual attribute analysis~\cite{farhadi2009objectattributes, lampert2009attributetransfer, berg2010attributediscovery, russakovsky2012attribute}.
%
% references using detection banks for high-level events (more related to video): mazloom2013conceptbanks,habibian2013eventcocneptvocabularies,mazloom2013queryingsignatures
%
It also has a simpler focus than image captioning which aims to
describe an image as completely and objectively as possible
\cite{ordonez2011im2text, karpathy2015}.

\labelnoise{tp}
\overview{bp}

ANP labeling is related to broader and more abstract image analysis for aesthetics~\cite{datta2006studying, marchesotti11aesthetic}, interestingness~\cite{isola11memorable}, affect or emotions~\cite{machajdik10affective, yanulevskaya08emotioncategorization, yanulevskaya12eyeofbeholder}.
Borth et al.~\cite{borth2013vso_sentibank} uses a bank of linear SVMs (SentiBank),  and \cite{chen2014deepsentibank} uses deep CNNs.  Both approaches aim to only detect known ANP from the dataset.  Deep CNNs have also been used for sentiment prediction ~\cite{xu2014visual, you2015robust},  but they are unable to model sentiment prediction by a mid-level representation such as ANPs.

Our goal is to map an image onto embedding derived from the visual sentiment ontology~\cite{borth2013vso_sentibank} that is built completely from visual 
data and respects visual correlations along adjective (A) and noun (N) semantics.  
By conditioning A on N, the combined concept of ANP becomes more visually detectable; by partitioning the visual space of nouns along adjectives, ANP forms a unique 
two-factor embedding for visual learning.

ANP images in \fig{anptable} exhibit structured correlations.  Along
each N column is the same type of objects; across the N columns are
related objects and parts.  Along each A row is the same type of
positive sentiment manifested in different objects; across the A rows
are sometimes interchangeable sentiments but most times distinctive
ones in their own ways.  For example, not every ANP is popular on
platforms like Flickr: {\it adorable eyes} and {\it attractive baby}
are not frequent enough to have associated images in the visual
sentiment dataset ~\cite{borth2013vso_sentibank}, suggesting that {\it
  adorable} is reserved more for overall impressions, whereas {\it
  attractive} is more for sexual appeal.  When an ANP classifier captures the 
row-column structure, it can fill in the semantic blanks where there is no
training data available and extend the concept consistent with other
known ANPs.

Learning a data-driven factorized adjective-noun embedding is
necessary not only for finding semantic structures, i.e., some ANPs
are more similar than others ({\it pretty girls} and {\it attractive
  girls} vs. {\it ugly girls}), but also for filtering out annotation
noise and removing inherent ambiguity.  \fig{labelnoise} illustrates
issues common to ANP images.  The same noun could mean different
entities: {\it baby} often refers to {\it human baby}, but it could
also refer to one's {\it pet} or {\it favorite thing} such as {\it
  cupcakes}, whereas an adjective could be used in a sarcastic manner
to indicate an opposite sentiment, and such usage is dependent on the
particular noun that it is paired with: images tagged as {\it
  attractive girls} are mostly positive, but images tagged as {\it
  attractive face} are often negative, with people making funny faces.

We present a nonlinear factorization model for ANP classification
based on the composition of two deep neural networks (\fig{overview}).  Unlike the
classical bilinear factorization model
\cite{freeman-tenenbaum:bilinear97} which decomposes an image into
style and content variations in a generative process, our model is
discriminative and nonlinear.  Compared to the bilinear SVM
classifiers \cite{pirsiavash:bilinearSVM09} which represents the
classifier as a product of two low-rank matrices, our model learns
both the feature and the classifier in a deep neural network
achitecture.  

\models{tp}

We emphasize that our factorized ANP CNN is only
seemingly similar to the recent bilinear CNN model
\cite{lin:bilinearCNN15}; we differ completely on the problem,
the architecture, and the technical approach.
{\bf 1)} The bilinear CNN model \cite{lin:bilinearCNN15} is a feature
extractor;  it takes the particular form of CNN products and the two CNNs
have no particular meaning.  Our bilinear model reflects
structured outputs and is in fact independent of how we extract the
features, i.e., we could additionally use their bilinear model for
feature extraction.  As a result, we can deal with unseen class labels,
while theirs does not address any such issue.
{\bf 2)} The blinear CNN model generalizes spatial pooling and only uses
conv layers, whereas the bilinear term of our model is a product of
latent representations for A and N, two aspects of the label, the
effect of which is entirely different from spatial pooling.

We explicitly map the output of A and N nets onto an individual
representation that is to be combined bilinearly for final
classification.  Such a factorization provides our model not only the much needed
regularization across different ANPs, but also the capability to 
classify and retrieve ANPs never seen during the
training.  Experimental results on the publicly available dataset
\cite{borth2013vso_sentibank}
%\cite{borth2013vso_sentibank}\footnote{\url{http://www.sentibank.org}} 
demonstrate that our model significantly outperforms 
independent ANP classification  on unseen ANPs, and on
retrieving images of new ANP vocabulary.  That is, our model based on
a factorized representation of ANP not only generalizes
better, but can also expands the ANP vocabulary on its own.

\section{Sentiment ANP CNN Classifiers}\label{sec:models}

We develop three CNN models that output a sentiment ANP label for an input image (\fig{models}). The first is a simple classification model that treats each ANP as an independent class, whereas the other two models have separate A and N processing streams and can thus predict ANPs never seen in the training data.  The second model is based on a shared-CNN architecture with two forked output layers, and the third model further incorporates an explicit factorization layer for A and N which is subsequently multiplied together for representing the ANP class.

{\bf ANP-Net: Basic ANP CNN Classifier.}  
\fig{models}a shows the baseline model that treats the ANP prediction as a straightforward classification problem.  We use VGG 16-layer ~\cite{simonyan2015vgg} as a base model and replace the final fully connected layer ``fc8'' from predicting 1,000 ImageNet categories to predicting 1,523 sentiment ANP classes.  The model is fine tuned from the ImageNet pretrained version.  We minimize the cross entropy loss with respect to the entire network through mini-batch stochastic gradient descent with momentum.  Given image $\mathbf{I}$ and ground truth label $t$, the cross entropy loss between $t$ and the softmax of $K$-category network output vector $\mathbf{y} \in \mathbb{R}^K$ is defined as \begin{align}\label{eq:xent}
                          {\cal L}(t, \mathbf{I}, \mathbf{\theta}) &= -\log p(y=t|\mathbf{I})\\
                          p(y=k|\mathbf{I}) & = {\rm softmax}(\mathbf{y})_k = \frac{\exp(\mathbf{y}_k)}{\sum_{m=1}^{K}\exp(\mathbf{y}_{m})} 
\end{align}
%~\cite{chen2014objectsentibank} which predict ANPs seen in training set directly on top of CNN by replacing the final output layer of AlexNet~\cite{krizhevsky2012imagenetnn} with ANP classification layer and finetuning whole network layer. Only difference from it is we use VGGNet as a base model (i.e. conv1-5 and fc6-7).

{\bf Fork-Net: Forked Adjective-Noun CNN Classifier.}
\fig{models}b shows an alternative model which predicts A and N separately from the input image.  The two streams share earlier layers of computation.  That is, the network tries to learn first a common representation useful for both A and N, and then an independent classifier for A and N separately.  As for ANP-Net, we use softmax cross-entropy loss for the A or N output, i.e., the network tries to learn universal A and N classifiers regardless of which N or A they are paired with, ignoring the correlation between A and N.
At the test time, we calculate the ANP response score from the product of A output $\mathbf{y}_A$ and N output $\mathbf{y}_N$:
\begin{align}
%    p(a=i|\mathbf{I}) = {\rm softmax}(a_i)
%    p(n=j|\mathbf{I}) = {\rm softmax}(n_j)
    p(y=(i,j)|\mathbf{I}) = p(y_A=i|\mathbf{I})\times p(y_N=j|\mathbf{I})
\end{align}

{\bf Fact-Net: Bilinearly Factorized ANP CNN Classifier.}  \fig{models}c shows a model with early layers of Fork-Net followed by a new product layer that combines the A and N outputs bilinearly for the final ANP output.  That is, with adjective $i$ and noun $j$ represented in the same $M$-dimensional latent space, $\mathbf{a}_i \in \mathbb{R}^M$ and $\mathbf{n}_j \in \mathbb{R}^M$ respectively, where $a_{im}$ and $n_{jm}$ denote $m$-th hidden variable for adjective $i$ and noun $j$, the Fact-Net output $y_{ij}$ is
$y_{ij} = \sum_{m\in M} a_{im} n_{jm}.$ Let $A,N$ denote the numbers of adjectives and nouns.  We have in matrix notations: \begin{align}
      \mathbf{Y}_{A\times N} &= \mathbf{A}_{A\times M} \cdot \mathbf{N}_{N\times M}', \\
\mathbf{A} &= \bmat{\mathbf{a}_{1}\\ \mathbf{a}_{2}\\ \vdots\\ \mathbf{a}_{A}}, \quad
     \mathbf{N} = \bmat{\mathbf{n}_{1}\\ \mathbf{n}_{2}\\ \vdots\\ \mathbf{n}_{N}}.
   \label{eq:factmat}
\end{align}
The Fact-Net learns  to map an image to a factorized A-N matrix representation $\mathbf{Y}$ by minimizing a cross entropy loss ${\cal L}$, with
gradients over latent A and N net outputs:
\begin{equation}\label{eq:factmatgrada}
\frac{\partial {\cal L}}{\partial \mathbf{A}} = \frac{\partial {\cal L}}{\partial \mathbf{Y}} \mathbf{N},\qquad
\frac{\partial {\cal L}}{\partial \mathbf{N}} = \left(\frac{\partial {\cal L}}{\partial \mathbf{Y}}\right)' \mathbf{A}.
\end{equation}
The entire network can be learned end-to-end with back propagation.  We find the network to learn better with the softmax function normalizing only over ANPs seen in the training set, in order to ignore the effect of ANP activations $Y_{ij}$ which are unseen during training. 

{\bf N-LSTM-A and A-LSTM-N} are two baseline recurrent algorithms, where networks predict ANPs sequentially. For example,~\fig{models}d first predicts the best noun given an image (i.e. $p(y_N=j|\mathbf{I})$), and then conditioned on the noun, an adjective is predicted $p(y_A=i| y_N=j ,\mathbf{I})$. Likewise,~\fig{models}e predicts first the best adjective, and then the best noun conditioned on that.
These two networks are inspired by image captioning models, most of which are in response to the creation of the MSCOCO Dataset~\cite{MScoco}.

\section{Experiments and Results}

We describe our ANP ontology and its associated publicly available dataset, present our experimental details, and show our detection and retrieval performance.

{\bf ANPs from Visual Sentiment Ontology (VSO).}  VSO was created by
mining online platforms such as Flickr and Youtube by the 24 emotions
from Plutchnik's Wheel of Emotions \cite{borth2013vso_sentibank}.
Derived from an analysis of tags associated with retrieved images and
videos from this mining process, an ontology of roughly 3,000 ANPs was
established,  e.g. {\it beautiful flowers} or {\it sad eyes},  See
Table \ref{tab:retrieval_overview} for statistics.

\vsotable{hbp}

{\bf ANP Dataset.}  We use the publicly available dataset of Flickr
images introduced in~\cite{borth2013vso_sentibank} with SentiBank
1.1. Please note that we experiment on the larger ``non-creative
common (Non-CC)'' also refered to as the ``Full VSO'' dataset and not
the smaller ``creative common (CC)'' only dataset.  In the Non-CC
dataset, for each ANP, at most 1,000 images tagged with the ANP
have been downloaded, resulting in about one million
images for the 3,316 ANPs of the VSO.

We first filter out the ANPs with fewer than 200 images, as such small
categories are either non-representative or with poorly generalizable
evaluation.  We also remove ANPs which have unintended semantics
against their general usage, e.g. {\it dark funeral} refers to images
of a heavy-metal band.  We then remove any ANP that have fewer than
two supports on both the adjectives and the noun, i.e. two ANPs
support each other if they share A or N.  Such pruning results
in 1,523 ANPs with $737,264$ images, $172$ adjectives and $240$
nouns. For each ANP, 20\% of images are randomly selected for testing,
while others are used for training.  We do ensure that images of one
ANP from an uploader (Flickr user) go to either training or
testing but not both, i.e., there is no user sharing between training
and testing images.

Our ANP labels come from Flickr user tags for images.  These labels
may be incomplete and noisy, i.e., not all true labels are annotated
and there could be falsely assigned labels. We do not manually refine
them; we use the labels as is and thus will refer to them {\it pseudo
  ground truth} (PGT).

{\bf Model Details.}  We fine tune the models in \fig{models}
from VGG-net pretrained on ImageNet dataset.  For ANP-net, the fully
connected layer for final classification is randomly initialized. For
Fork-net and Fact-net, we initialize {\it fc7-a} and {\it fc7-n} and all the
following layers randomly. The {\it fc7-a} and {\it fc7-n} layers are followed
by a parametric ReLU (PReLU) layer for better convergence
\cite{kaiming2015prelu}. We use 0.01 for the learning rate throughout
training except the learning rates of pretrained weights are reduced
by a factor of 10.  Our models are implemented using our modified
branch of CAFFE \cite{jia2014caffe}. We use the polynomial decay
learning rate scheduler.  We train our models though stochastic
gradient descent with momentum $0.9$, weight decay $0.0005$ and
mini-batch size $256$ for five epochs, taking $2$-$3$ days for
training convergence on a single GPU.
For the two recurrent models, we expand and modify Andrej Karpathy's "neuraltalk" Github branch ~\cite{karpathy2015}. After incorporating various hidden layer sizes, we settle on a hidden layer of 128, word+image encoding size of 128, and a single recurrent layer. However, we did not pretrain on any word/semantic data on other corporat (e.g., MSCOCO), but rather only sequentially trained adjectives and nouns from Sentibank.

{\bf Top-$k$ Accuracy on Seen and Unseen ANPs.}  
ANP classes are either seen or unseen depending on whether the ANP
concept was given during training.  While images of an explicitly {\it
  unseen ANP} class, e.g. {\it beautiful men}, might be new to a
model, images sharing the same A or N, e.g. {\it beautiful girls} or
{\it handsome men}, have been seen by the model.  Our unseen ANPs come
from those valid ANPs which are excluded from training due to their
fewer than $200$ examples. For the unseen dataset for our evaluation,
we filter out the unseen ANPs with less than $100$ examples.  We have
$293$ unseen ANPs with $43,133$ examples in total.

We use top-$k$ accuracy, $k=1,5,10$, to evaluate a model.  We examine whether the PGT label of an image is among the top $k$ ANP labels suggested by a model output.  The average hit rate for test images of an ANP indicates how accurate a model is at differentiating the ANP from others.  The top-$k$ accuracy on seen ANPs shows how good the model is fitting the training data, whereas that on unseen ANPs shows how well the model can generalize to new ANPs.

We take the DeepSentiBank model \cite{chen2014deepsentibank} as a
baseline, which already outperforms the initial SentiBank 1.1. model
~\cite{borth2013vso_sentibank}.  It uses the AlexNet
architecture~\cite{krizhevsky2012imagenetnn} but fine-tuned to the ANP
classes.  We apply the same CNN architecture and setup to our set of
1,523 ANPs from the NON-CC dataset.  

\topkaccu{tp}

Table~\ref{tab:topk}a shows top-$k$ accuracy on \textit{seen}
ANPs. ANP-Net produces the best accuracies, since it is trained for
directly optimizing the classification accuracies on these
ANPs. Fact-Net outperforms Fork-Net for a number of choices of $M$,
suggesting that our factorized representation better captures
discriminative information between ANPs. Also, our VGG-net based
models all outperform the Alex-net based DeepSentiBank model,
confirming that deeper CNN architectures build stronger classification
models.

Table~\ref{tab:topk}b shows top-$k$ accuracy on \textit{unseen} ANPs.
Consistent with the results for \textit{seen} ANPs, Fact-Net always
outperforms Fork-Net.  More importantly, the top-$k$ accuracies on
unseen ANPs decrease with increasing $M$, with Fact-Net at $M\!=\!2$
significantly outperforms Fork-Net.  That is, the larger the internal
representation for A and N, the poorer the generalization to new
ANPs. Since models like DeepSentiBank or the individual ANP-net are
only capable of classifying what they have seen during training, we leave
the corresponding entries in the Table blank.

The top-$k$ accuracies on our two baseline image captioning models,
while significantly above the chance level due to the large number of
ANP classes, still seem surprisingly low.  These poor results
demonstrate the challenge of our ANP task, and in turn corroborate the
effectiveness of factorized ANP CNN model.

Our ANP task differs from the common language+vision problems in two
significant ways: {\bf 1)} It aims to capture not so much the
semantics of word adjective-noun pairs such as ({\it bull shark, blue
  shark, tiger shark}), but rather pairs of adjectives and nouns with
the sentiment evoked by pure visual experience such as ({\it cute dog,
  scary dog, dirty dog}). In this sense, our adjectives just happen to
be words, the subjective aspect of our labels for conditioning our
nouns, in order to partition the visual space instead of the semantic
space. Word semantics reflected in the co-occurrence of natural text
has little to do with our visual sentiment analysis. Our task is thus
entirely different from the slew of language model and image
captioning works.  {\bf 2)} We are generalizing not along a conceptual
hierarchy with obvious visual similarity basis, e.g. from boxer and
dog to canine and animal, but across two different conceptual trees
with subtle visual basis, e.g. from ({\it beautiful + sky / landscape /
person}) and ({\it dead / dry / .../ old + tree}) to ({\it beautiful tree}). Our
task is thus much more challenging.

\accugap{tp}

{\bf Best and worst ANPs by Fact-Net.}
We look into the classification accuracy on individual ANPs and
compare our Fact-Net with $M=2$ against the best alternative ANP-Net
for the seen ANPs and Fork-Net for the unseen ANPs.  The former could
help us understand how the training data are organized and the latter
how the model could fill in the blanks of the ANP space and generalize
to new classes.

\retrievesource{!tp}

\fig{accugap}a lists the top and bottom 10 seen ANPs when they are
sorted by the difference in top-10 accuracy between Fact-Net and
ANP-Net, and \fig{accugap}b lists the top and bottom 10 unseen ANPs
when they are sorted by the difference in top-10 accuracy between
Fact-Net and Fork-Net.  The range of accuracy gap is $(-0.6,0.4)$ for
the unseen, much wider than $(-0.3,0.3)$ for the seen ANP case.  We
analyze the accuracies with respect to the number of images as well as
the number of different ANPs seen during the training, and obtain
correlation coefficients at the order of 0.05, suggesting that the
gain of individual ANPs cannot be explained by the amount of exposure
to training instances, but it has more to do with the correlations
between ANPs.  \fig{accugap}c-d show sample images from the top and
bottom 3 ANPs for the seen and unseen ANPs.  Our Fact-Net always seems
to have a larger gain over ANPs with fewer varieties.

{\bf Image Retrieval by Fact-Net and Fork-Net.}
We also compare models on retrieving images of a particular ANP.  We
rank the model output for all the images corresponding to the ANP, and
return the images with top responses.  The ANP could be seen or unseen
in our dataset, or completely novel, e.g. {\it dangerous summer}.  
%For ANP-net, there is no representation for unseen or novel ANPs, and
%we use the corresponding output component for the seen ANP to rank
%the images; 
For Fork-net, we use the product of A-net and N-net output
components corresponding to the ANP parts; for Fact-net, we use the
output component directly corresponding to the ANP.

\fig{retrieve} shows side-by-side comparisons of top retrievals for 1
seen ANP ({\it beautiful sky}) and 3 unseen/novel ANPs by Fact-Net and
Fork-Net.  {\bf 1)} Images returned by Fact-Net are in general more
accurate on the noun: e.g. {\it ugly baby} and {\it ugly sky} images
contain mostly baby and sky scenes, whereas Fork-Net results contain
mostly fish and buildings instead.  {\bf 2)} Fact-Net retrievals have
more varieties on the adjective: e.g. {\it beautiful sky} images have
both warm and cool color tones, whereas Fork-Net results have mostly
cool color tones.  {\bf 3)} Fact-Net results correct more annotation
mistakes: e.g. man with a tie tagged {\it hot girls} is rightly
retrieved for {\it beautiful men}, whereas those mistakes such as
females tagged {\it sexy fashion} and {\it fragile body} are retained
in Fork-Net results for {\it beautiful men}. {\bf 4)} Our Fact-Net can
also be used for consensus re-tagging: while beauty is in the eyes of
the beholder, we see that the images tagged {\it beautiful sky} become
top retrievals for {\it ugly sky}, which do share characteristics with
other gloomy scenes.

%\section{Conclusions}
{\bf Conclusions.}
From our extensive experimentation, we gain two insights into the
unique and challenging sentiment ANP detection task, unlike other well-defined
image classification or captioning task.  {\bf 1)} For seen ANPs, the
ANP-net is the winner, but it cannot handle unseen ANPs, a killing
caveat.  We set our initial goal to exceed the ANP-net baseline, after
numerous trials, we realize that there will always be a baseline
version that no factorized model could beat, since the former directly
optimizes the performance over each ANP.  However, such CNNs neither
see the connections between A’s and N’s nor generalize as ours.  {\bf
  2)} Our Fact-Net on unseen ANPs is substantially better than all
baselines.  In addition, due to noisy labels (\fig{labelnoise} and
\fig{retrieve}), the results are actually even better: e.g., in
\fig{retrieve}, {\it beautiful men} retrieves correct results with
wrong user labels of {\it hot girls} or {\it cold beer}.  Our
factorized ANP CNN not only trains better from noisy labels,
generalizes better to new images, but can also expands the ANP
vocabulary on its own.

{\footnotesize
\bibliographystyle{plain}
\bibliography{anp.bib}
}

\end{document}